\documentclass[3p,times]{elsarticle}

\usepackage{amsmath}
\usepackage{amssymb}
\usepackage{hyperref}
\usepackage{multirow} 

\begin{document}

\begin{frontmatter}

\title{A Foundation Model Approach for Fetal Stress Prediction During Labor From  cardiotocography (CTG) recordings}

\author[1]{Naomi Fridman}
\ead{naominoe.fridman@gmail.com}

\author[2]{Berta Ben Shachar}

\address[1]{Department of Industrial Engineering, Ariel University, Ariel, Israel}
\address[2]{NF Algorithms \& AI, Tel Aviv, Israel}


\begin{abstract}
Intrapartum cardiotocography (CTG) is widely used for fetal monitoring during labor, yet its interpretation suffers from high inter-observer variability and limited predictive accuracy. Deep learning approaches have been constrained by the scarcity of CTG recordings, particularly those with clinical outcome labels. We present the first application of self-supervised pre-training to intrapartum CTG analysis. Our approach leverages 2,444 hours of unlabeled recordings for masked pre-training, followed by fine-tuning on the 552-recording CTU-UHB benchmark. We employ a PatchTST transformer architecture with a channel-asymmetric masking scheme designed for fetal heart rate (FHR) signal reconstruction. We achieve an area under the receiver operating characteristic curve (AUC) of 0.83 on the full test set and 0.853 on uncomplicated vaginal deliveries, exceeding previously reported results on this benchmark (AUC 0.68--0.75). Error analysis reveals that false-positive alerts typically correspond to CTG patterns judged concerning on retrospective clinical review, suggesting clinically meaningful predictions even when umbilical pH is normal. We release standardized dataset splits and model weights to enable reproducible benchmarking. Our results demonstrate that self-supervised pre-training can address data scarcity in fetal monitoring, offering a path toward reliable decision support in the labor room.
\end{abstract}

\end{frontmatter}

\section{Introduction}
\subsection{Clinical Background}
Annually, approximately two million babies are stillborn worldwide, with over 40\% of these events occurring during labor \cite{unigme2023}. Many of these deaths are preventable through high-quality monitoring during pregnancy \cite{vogel2014} and timely obstetric intervention \cite{bhutta2014}. Intrapartum asphyxia is a major contributor to stillbirths \cite{goldenberg2016}, which occurs when inadequate oxygen supply to the fetus during labor leads to hypoxia. The fetal oxygen supply relies entirely on uteroplacental blood flow across the umbilical cord; any disruption—whether from uterine contractions temporarily compressing placental vessels or umbilical cord compression—can reduce oxygen delivery to fetal tissues \cite{ayres2015figo}. During normal labor, uterine contractions cause transient reductions in FHR (decelerations), from which a healthy fetus rapidly recovers. Failure to recover, prolonged decelerations, or reduced variability can indicate fetal compromise \cite{bennet2009}. Since these responses manifest as characteristic FHR patterns, continuous monitoring during labor has become standard practice for identifying fetal compromise \cite{ayres2015figo}.

\begin{figure}[t]
\centering
\includegraphics[width=\textwidth]{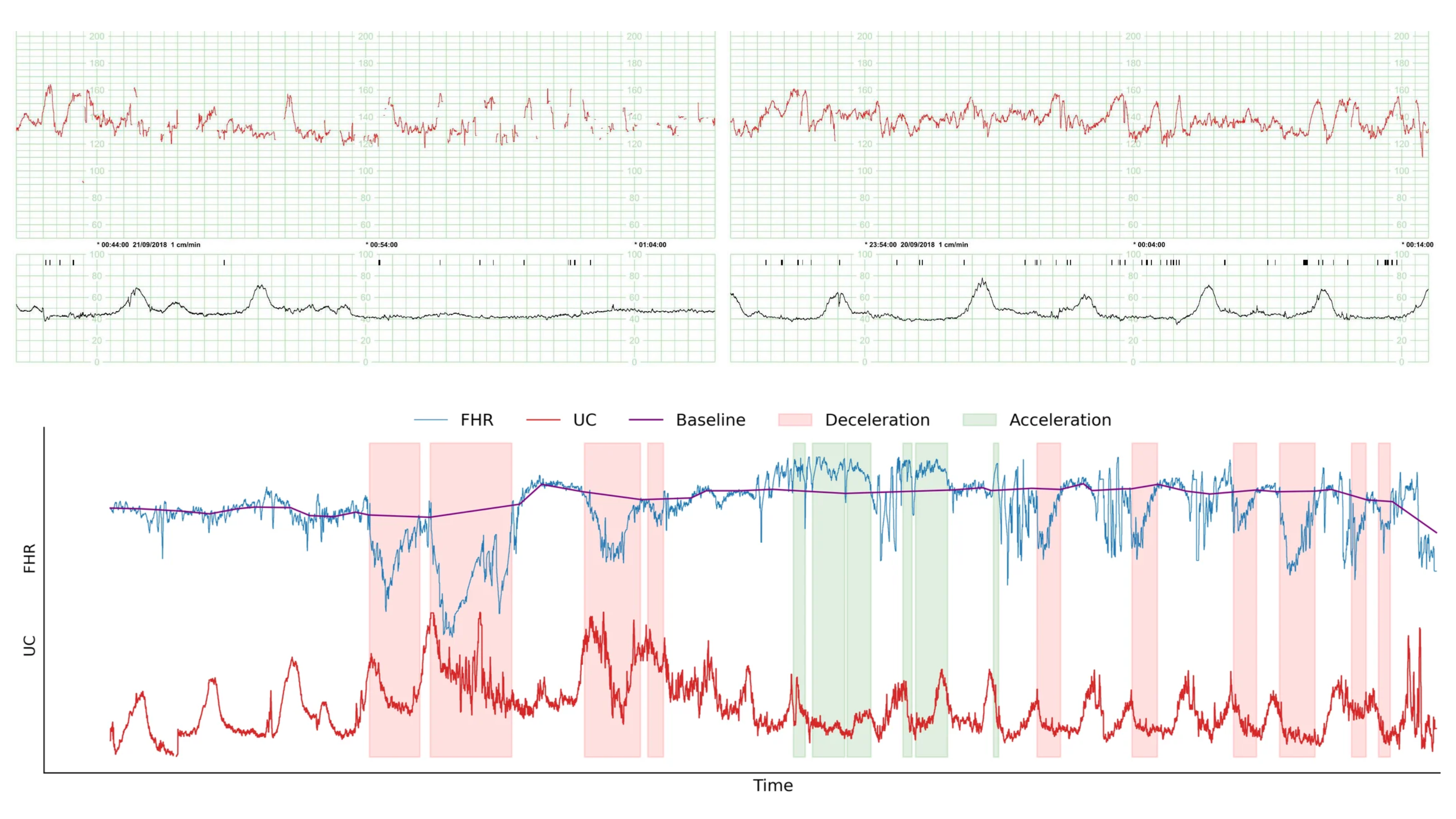}
\caption{Cardiotocography (CTG) recording examples. \textbf{Top:} Clinical CTG display as seen by clinicians on bedside monitors, showing fetal heart rate (FHR, red, upper panel) and uterine contractions (UC, black, lower panel) on standard grid paper (1 cm = 1 min). Visual interpretation requires simultaneously tracking baseline, variability, accelerations, decelerations, and their temporal relationship to contractions. \textbf{Bottom:} Annotated CTG recording from the FHRMA dataset \cite{boudet2019fhrma}, illustrating expert consensus annotations of the FHR baseline (purple), accelerations (green shading), and decelerations (pink shading). Even with explicit annotations, classification of these features shows substantial inter-observer variability in clinical practice.}
\label{fig:ctg_example}
\end{figure}

Intrapartum cardiotocography (CTG) records FHR alongside uterine contractions (UC), enabling clinicians to assess fetal wellbeing throughout labor (Figure~\ref{fig:ctg_example}, top). Despite widespread adoption, CTG interpretation remains problematic due to high false-positive rates \cite{alfirevic2017}, subjectivity in visual interpretation \cite{bernardes1997, palomaki2006}, and substantial inter-observer variability \cite{schiermeier2011}. These challenges are exacerbated in low-resource settings \cite{blencowe2016, lawn2016}.

The FIGO guidelines define standardized morphological features for CTG interpretation: baseline FHR, variability, accelerations, and decelerations (Figure~\ref{fig:ctg_example}, bottom) \cite{ayres2015figo}. However, these criteria assume artifact-free signals rarely achieved in practice. Visual interpretation requires tracking multiple features simultaneously while correlating FHR changes with contraction timing—a demanding task given that monitors display only a limited time window of the recording.

Model development is further complicated by the lack of a definitive outcome measure. Umbilical artery pH offers an objective biochemical marker but does not fully capture long-term neurological outcome, while Apgar scores are subjectively assigned and influenced by factors beyond intrapartum events. Moreover, retrospective CTG interpretation by experts has been shown to be biased by knowledge of delivery outcome \cite{ayres2011bias}.

\subsection{Technical Background}
Machine learning approaches to CTG analysis have evolved through several methodological paradigms. Early computational methods applied classical signal processing techniques to FHR signals, including spectral analysis, time-frequency decomposition, and nonlinear dynamics measures such as entropy and fractal dimension \cite{warrick2010, warrick2012}. These extracted features were then used with traditional classifiers—support vector machines, random forests, or logistic regression \cite{hoodbhoy2019, pradhan2021}. Automated systems such as SisPorto \cite{sisporto2000} demonstrated the feasibility of computerized CTG interpretation using rule-based analysis of FIGO-defined morphological features. Ben M'Barek et al. \cite{mbarek2023} achieved AUC of 0.74 on CTU-UHB using logistic regression with such FIGO-based features.

A parallel line of work leveraged signal processing to transform one-dimensional CTG signals into two-dimensional representations suitable for image classification networks. DeepFHR \cite{zhao2019} applied continuous wavelet transforms to convert FHR signals into time-frequency scalograms, then classified these images with 2D CNNs. Similar approaches have used spectrograms, recurrence plots, and histogram-based representations as inputs to standard image architectures.

End-to-end deep learning methods process raw or minimally preprocessed signals directly. One-dimensional CNNs and recurrent architectures (LSTMs, GRUs) learn representations from the signal itself without explicit feature engineering. CTGNet \cite{fergus2020} employed 1D CNNs with depthwise separable convolutions to process FHR and UC signals jointly. Ogasawara et al. \cite{ogasawara2021} proposed CTG-net, achieving AUC of 0.73 on their institutional dataset and 0.68 on CTU-UHB. Petrozziello et al. \cite{petrozziello2019} applied multimodal CNNs to over 35,000 births, reporting AUC of 0.81 on their Oxford dataset and approximately 0.75 on CTU-UHB.

However, direct comparison across studies is challenging due to heterogeneous outcome definitions—some use umbilical artery pH with varying thresholds (< 7.05, 7.10, 7.15, or 7.20), others use Apgar scores, and some combine multiple criteria. Additionally, many high-performing models are trained and tested on private institutional datasets, preventing independent validation. Reported AUC values range from 0.68 to over 0.95, but the highest figures typically reflect evaluation on private data or methodological issues. The public CTU-UHB database \cite{chudacek2014} remains the only substantial benchmark; on this dataset, properly evaluated methods achieve AUC of 0.68--0.75.

Beyond these methodological inconsistencies, there are also conceptual reasons to question near-perfect classification results. Deep learning models are typically trained on short CTG segments (3--10 minutes) extracted from longer recordings (60--90 minutes), with each segment labeled according to the final umbilical artery pH. However, fetal compromise is a dynamic process—pathological patterns may only manifest during specific portions of the recording, not throughout its entirety. Labeling every segment from an acidemic delivery as positive assumes the abnormality is continuously detectable, which is clinically unrealistic, and likely inflates performance estimates on short segments.

Transformer architectures have revolutionized sequential modeling across domains \cite{kalyan2021, khan2021, karita2019}. Their self-attention mechanism enables learning long-range dependencies without the limitations of recurrent architectures. PatchTST \cite{patchtst2023} introduced patch-based tokenization for time series, improving computational efficiency and representational capacity. Applications of transformers to CTG remain limited: Wu et al. \cite{wu2024etcnn} proposed ETCNN, a hybrid transformer-CNN for morphological feature detection (baseline, accelerations, decelerations), but no prior work has applied transformers to intrapartum CTG outcome classification or explored self-supervised pre-training for CTG.

In this work, we adopt the foundation model paradigm for CTG analysis: a high-capacity model is first pre-trained to reconstruct masked fetal heart rate (FHR) segments, thereby learning general representations of CTG signals in a self-supervised manner, and is only subsequently fine-tuned for specific downstream clinical tasks such as fetal compromise classification. Masked pre-training improves representation learning on any available data; using unlabeled recordings further enlarges the effective training corpus.

Concretely, we apply self-supervised masked pre-training to intrapartum CTG using the CTGDL dataset \cite{ctgdl2025}, which integrates CTU-UHB with two additional CTG databases—FHRMA (135 recordings) \cite{boudet2019fhrma} and SPaM'17 intrapartum CTG challenge database (297 recordings) \cite{spam2017challenge,georgieva2019computer}—for a total of 2,444 hours of fetal monitoring data. These additional recordings lack clinical outcome labels and have therefore not been used in prior CTU-UHB benchmarking studies. We base our architecture on PatchTST \cite{patchtst2023}, extending it with channel-asymmetric masking tailored to the distinct clinical roles of FHR and UC signals. For comparison, Khan et al. \cite{khan2024patchctg} applied standard PatchTST to antepartum CTG classification, achieving AUC of 0.77 on a private dataset of over 20,000 labeled recordings. Using the same base architecture with our masked pre-training approach and channel-asymmetric masking, we achieve AUC of 0.83 on the full CTU-UHB test set and 0.853 on vaginal deliveries—demonstrating that self-supervised pre-training combined with task-specific architectural modifications can outperform fully supervised training on substantially larger datasets.

\section{Methods}

\subsection{Dataset}
We utilize the CTGDL dataset \cite{ctgdl2025}, which integrates three complementary CTG databases totaling 984 recordings and 2,444 hours of fetal monitoring data. The dataset comprises: CTGDL\_CTU\_UHB (552 recordings with clinical outcomes including umbilical artery pH) \cite{chudacek2014open}, CTGDL\_FHRMA (135 recordings with expert morphological annotations) \cite{boudet2019fhrma}, and CTGDL\_SPAM (297 long-duration recordings from the CTG Challenge 2017). All recordings contain dual-channel signals sampled at 4 Hz: fetal heart rate (FHR) and uterine contractions (UC).

The complete dataset was used for self-supervised pre-training via masked prediction. For the downstream classification task, we used the CTGDL\_CTU\_UHB subset, which originates from the Czech Technical University and University Hospital Brno Intrapartum Cardiotocography Database \cite{chudacek2014open, physionet2000}. This database contains carefully selected recordings collected between 2010 and 2012 using Philips Avalon FM50 monitors. Selection criteria included: singleton pregnancy, gestational age >36 weeks, no known developmental defects, second stage labor duration $\leq$30 minutes, and FHR signal quality >50\% in each 30-minute window. Each recording starts no more than 90 minutes before delivery (mean duration 74.2 $\pm$ 7.6 minutes). Clinical characteristics are summarized in Table~\ref{tab:clinical}.

The dataset includes both uncomplicated and high-risk deliveries, with documented cases of cesarean delivery (8.3\%), induced labor (39.3\%), meconium-stained fluid (11.6\%), labor arrest (10.0\%), hypertension (8.0\%), and diabetes (6.7\%). These rates exceed typical population averages, which is advantageous for studying the critical cases where CTG monitoring is most essential, but presents a challenge for developing models that generalize to the broader, lower-risk obstetric population.

\begin{table}[h]
\centering
\caption{Clinical characteristics of the CTGDL\_CTU\_UHB dataset (n=552)}
\label{tab:clinical}
\begin{tabular}{llc}
\hline
\textbf{Category} & \textbf{Parameter} & \textbf{Value} \\
\hline
Maternal & Age (years) & 29.7 $\pm$ 4.5 \\
\hline
\multirow{3}{*}{Delivery} & Vaginal & 506 (91.7\%) \\
 & Cesarean section & 46 (8.3\%) \\
 & Gestational age (weeks) & 40.0 $\pm$ 1.1 \\
\hline
\multirow{4}{*}{Neonatal} & Birth weight (g) & 3400 $\pm$ 455 \\
 & Sex (Male/Female) & 286/266 \\
 & Apgar 1 min & 9 [8--9] \\
 & Apgar 5 min & 9 [9--10] \\
\hline
\multirow{2}{*}{Outcome} & Umbilical artery pH & 7.23 $\pm$ 0.10 \\
 & pH < 7.15 (Acidemia) & 113 (20.5\%) \\
\hline
\multirow{8}{*}{Risk factors} & Induced labor & 217 (39.3\%) \\
 & Abnormal amniotic fluid & 147 (26.6\%) \\
 & Meconium-stained fluid & 64 (11.6\%) \\
 & Labor arrest (no progress) & 55 (10.0\%) \\
 & Non-cephalic presentation & 53 (9.6\%) \\
 & Hypertension & 44 (8.0\%) \\
 & Diabetes & 37 (6.7\%) \\
 & Preeclampsia & 17 (3.1\%) \\
\hline
\end{tabular}
\\[0.5em]
{\small Values are mean $\pm$ SD, n (\%), or median [IQR].}
\end{table}

For binary classification, fetal acidemia was defined as umbilical artery pH < 7.15, yielding 113 positive cases (20.5\%). The dataset was split into training, validation, and test sets with stratified sampling to maintain consistent class distribution across splits (Table~\ref{tab:splits}). We publish these splits to establish a reproducible benchmark for future studies.

\begin{table}[h]
\centering
\caption{Dataset splits for reproducibility}
\label{tab:splits}
\begin{tabular}{lccc}
\hline
\textbf{Split} & \textbf{n} & \textbf{Acidemia (pH<7.15)} & \textbf{Prevalence} \\
\hline
Train & 441 & 90 & 20.4\% \\
Validation & 56 & 12 & 21.4\% \\
Test & 55 & 11 & 20.0\% \\
\hline
Total & 552 & 113 & 20.5\% \\
\hline
\end{tabular}
\end{table}

\subsection{Signal Preprocessing}
Signal preprocessing followed the CTGDL pipeline \cite{ctgdl2025}: physiologically implausible FHR values (<50 or >220 bpm) were removed, brief artifacts  and transient spikes  were eliminated, and missing values were filled using linear interpolation.

\begin{figure}[h]
\centering
\includegraphics[width=\textwidth]{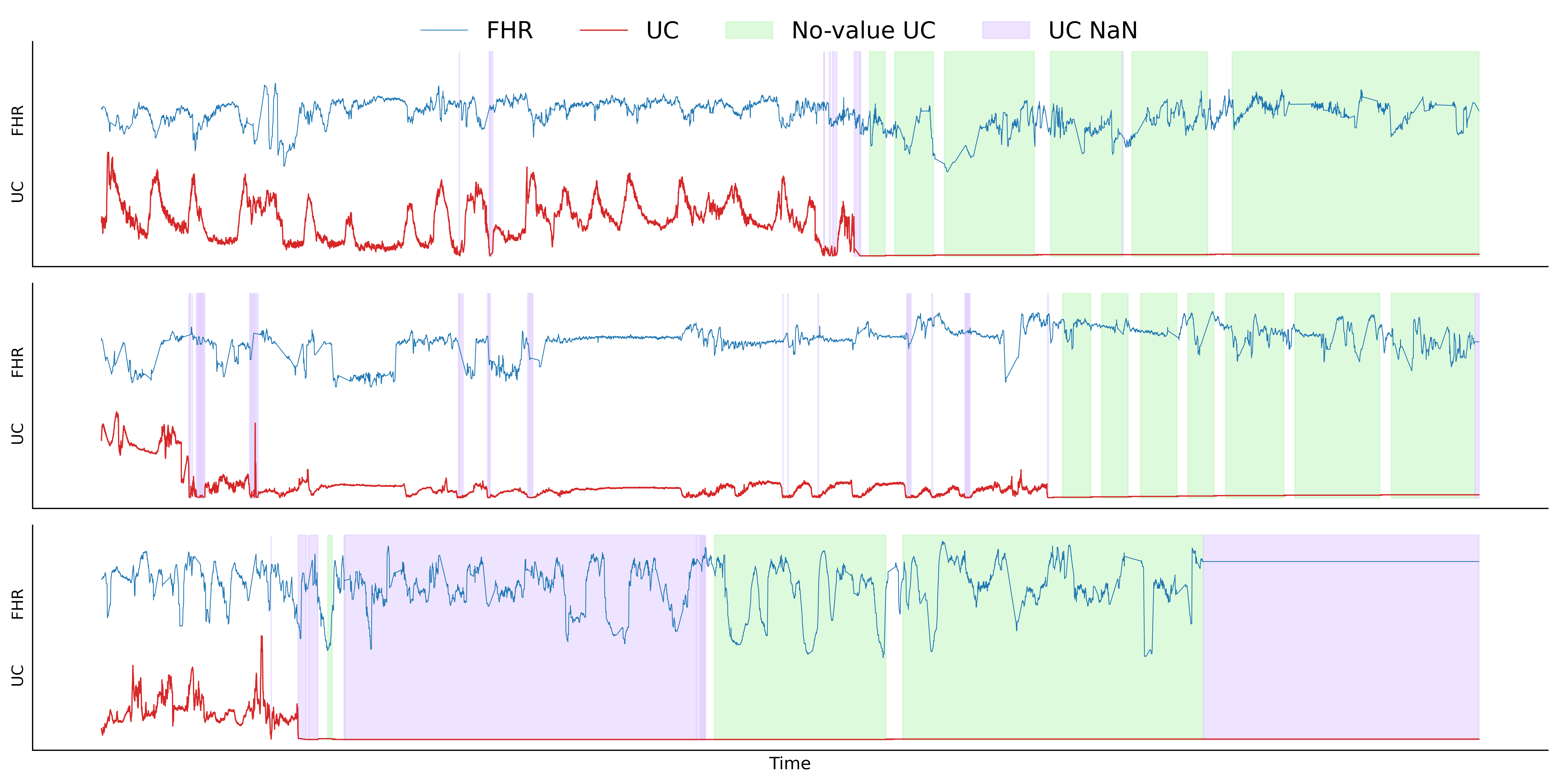}
\caption{Examples of UC signal quality issues in CTG recordings. Green shaded regions indicate detected no-value segments where the signal appears flat at non-zero levels due to sensor displacement. Purple shaded regions indicate original NaN values. These flat segments, while numerically valid, contain no physiological information and were excluded from analysis.}
\label{fig:uc_quality}
\end{figure}

We extended this pipeline to address a characteristic artifact in UC signals. Unlike FHR, where missing data appears as zero or NaN values, UC signals frequently contain flat regions at non-zero levels that represent sensor displacement or loss of contact rather than true physiological measurements. To identify such regions, we computed the rolling standard deviation over 30-second windows (120 samples at 4 Hz) and marked segments as invalid where the standard deviation fell below $10^{-5}$, the original value was not already flagged as missing, and the amplitude remained below 80 mmHg. This threshold was chosen empirically based on visual inspection of typical flat artifacts (Figure~\ref{fig:uc_quality}) illustrates this phenomenon across representative recordings.

For model input, signals were normalized to the [0,1] range. FHR values were clipped to 50--210 bpm and divided by 160. UC values were clipped to 0--100 mmHg and divided by 100.

\subsection{Model Structure}

We employ PatchTST \cite{patchtst2023}, a Transformer-based architecture designed for time series representation learning. Two key design principles make this architecture suitable for CTG analysis: patch-based tokenization and channel-independent processing.

\paragraph{Patch-Based Tokenization.}
The model segments each input signal into overlapping patches rather than processing individual time points. This approach captures local morphological structure within each patch—critical for CTG analysis where clinically relevant patterns such as accelerations and decelerations are defined by their shape over temporal windows rather than instantaneous values. Additionally, patch-based segmentation reduces the sequence length presented to the attention mechanism, enabling the model to capture relationships between distant events such as FHR responses to uterine contractions. Each patch is projected into a latent space through a learned linear transformation $W_P$, and learnable positional encodings $W_{pos}$ are added to preserve temporal ordering:
\begin{equation}
    x_d = W_P \cdot x_p + W_{pos}
\end{equation}
This formulation enables the self-attention mechanism to attend to patches based on both content similarity and relative temporal position—essential for detecting patterns like late decelerations, which are defined by their temporal relationship to uterine contractions.

\paragraph{Channel-Independent Processing.}
A key architectural choice leverages PatchTST's channel-independence design. Unlike channel-mixing approaches where all input features are projected into a shared  embedding space, channel-independence processes each univariate signal through the same Transformer backbone while maintaining separate forward passes. This design offers three advantages for CTG analysis: (1) each signal learns attention patterns tailored to its characteristics—FHR signals exhibit different temporal dynamics than UC signals; (2) shared weights act as an implicit regularizer, reducing overfitting on limited clinical data; and (3) the model remains robust to noise in one channel without propagating artifacts to the other. The embedded patches from both channels are processed through identical encoder layers—comprising multi-head self-attention and feed-forward networks with residual connections and batch normalization—but maintain independent attention computations. For downstream tasks, the encoded representations from both channels are concatenated, allowing the task-specific head to jointly consider learned features from both FHR and UC signals while preserving their independently learned attention patterns.  
\begin{figure}[t]
\centering
\includegraphics[width=\textwidth]{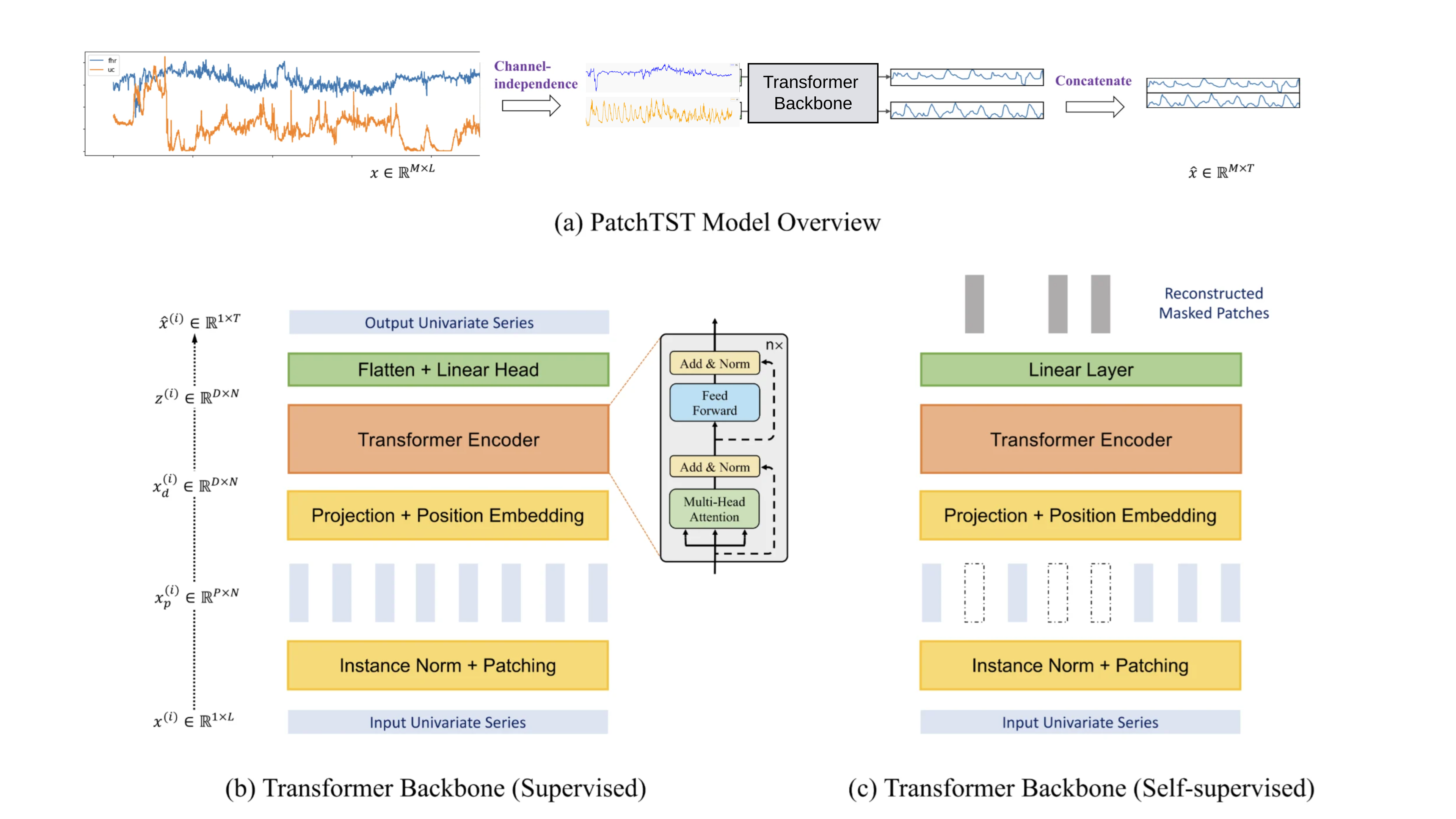}
\caption{PatchTST architecture overview, adapted from \cite{patchtst2023}. (a) Channel-independent processing: each channel (FHR, UC) is processed separately through the transformer backbone and concatenated for prediction. (b) Supervised learning: input series is normalized, segmented into patches, projected with positional embeddings, processed by the transformer encoder, and mapped to output via a linear head. (c) Self-supervised pre-training: random patches are masked and the model learns to reconstruct them, enabling representation learning from unlabeled data.}
\label{fig:patchtst}
\end{figure}

\subsection{Masked Pre-training}

\begin{figure}[ht]
    \centering
    \includegraphics[width=\textwidth]{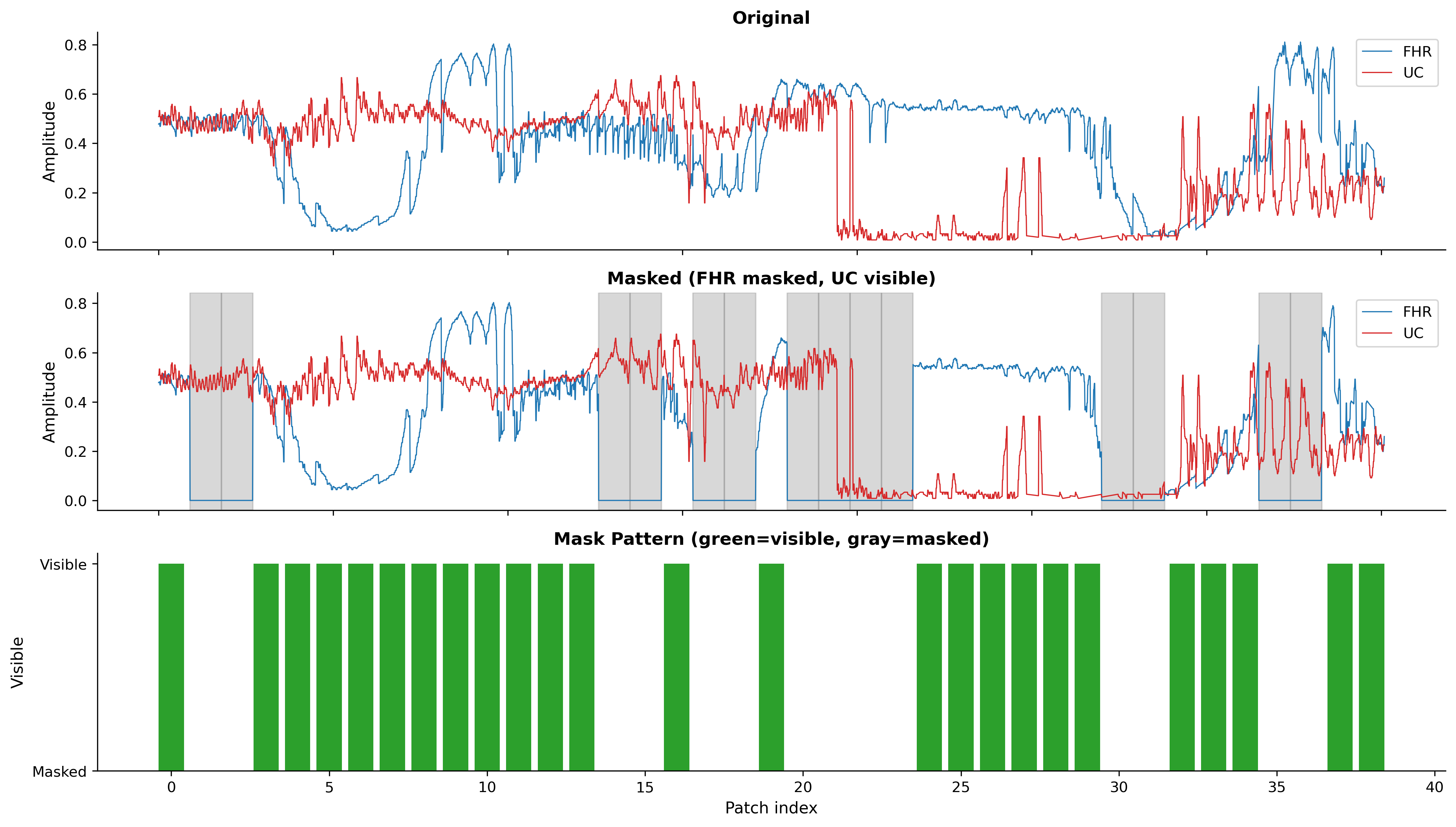}
    \caption{Illustration of the channel-asymmetric masking strategy. Top: original CTG signal with FHR (blue) and UC (red) channels. Middle: masked input where selected FHR patches are zeroed (gray regions) while UC remains intact. Bottom: binary mask pattern indicating masked (gray) and unmasked (green) FHR patches. The first and last patches are never masked, and masked patches appear in contiguous groups.}
    \label{fig:masking}
\end{figure}

For self-supervised pre-training, we adopt a masked autoencoder paradigm with a channel-asymmetric masking strategy. At each temporal position, the FHR and UC signals form a paired patch. During masking, only the FHR patch is replaced with zeros while the corresponding UC patch remains intact (Figure~\ref{fig:masking}). This asymmetric design forces the model to reconstruct occluded FHR segments using two complementary sources of information: (1) the surrounding FHR context, and (2) the temporally aligned UC signal. Consequently, the model learns to associate FHR patterns with concurrent uterine activity—capturing clinically relevant relationships such as FHR decelerations in response to contractions.

To prevent trivial reconstruction through simple interpolation from neighboring patches, we impose two constraints on the masking pattern. First, masked patches must appear in contiguous groups of at least two—isolated masked patches are not permitted. Second, the first and last patches of each sequence are excluded from masking to preserve boundary context. Approximately 40\% of the remaining patches are randomly selected for masking in each training sample.

The model is trained to reconstruct the original FHR values at masked positions using mean squared error:
\begin{equation}
    \mathcal{L}_{pretrain} = \frac{1}{|M|} \sum_{i \in M} \| x_i^{FHR} - \hat{x}_i^{FHR} \|_2^2
\end{equation}
where $M$ denotes the set of masked patch indices. This objective compels the model to learn meaningful representations of CTG patterns by inferring missing FHR segments from both surrounding temporal context and the concurrent UC signal.

Dropout in both the encoder and head was set to 0.2, matching the configuration reported for PatchTST and providing regularization that is effective for patch-based masked pretraining without substantially slowing convergence \cite{patchtst2023}. With a context length of 1{,}800 samples and a patching scheme of length 48 and stride 24, each segment is decomposed into 73 overlapping patches, which yields a sufficiently rich token sequence for masked modeling while keeping sequence length and memory usage manageable \cite{patchtst2023}. The masking ratio was fixed at 0.4, in line with published masked pretraining setups for PatchTST where a random mask ratio of 0.4 is commonly used, ensuring that the reconstruction task remains challenging enough to drive useful representation learning but not so extreme as to destabilize training \cite{patchtst2023}. Optimization was performed with the Adam optimizer and a learning rate of \(1\times10^{-4}\), a conservative value within the typical \(10^{-4}\)–\(10^{-3}\) range for Transformer-based pretraining on biomedical time series, chosen to favor stable convergence on a limited number of subjects over more aggressive but potentially noisy updates \cite{patchtst2023,chudacek2014ctuuhb}.

\subsection{Fine-tuning for Classification}
Following self-supervised pre-training, we fine-tuned the model for binary classification of fetal compromise, defined as umbilical artery pH < 7.15 (Figure~\ref{fig:patchtst}). The transformer encoder weights were initialized from the pre-trained backbone, while the forecasting head was replaced with a classification head consisting of a linear layer mapping the flattened encoder output to two classes.

To address class imbalance, we augmented the training set with cesarean delivery cases from the CTGDL\_SPAM dataset. These recordings, identified by stage 2 duration of zero, were labeled as positive cases under the assumption that intrapartum cesarean delivery typically indicates CTG abnormalities prompting intervention. This augmentation increased the representation of pathological patterns in the training data.

The model was fine-tuned for 100 epochs using AdamW optimizer with cross-entropy loss and early stopping based on validation AUC.

\subsection{Inference and Alert Generation}
\begin{figure}[t]
\centering
\includegraphics[width=\textwidth]{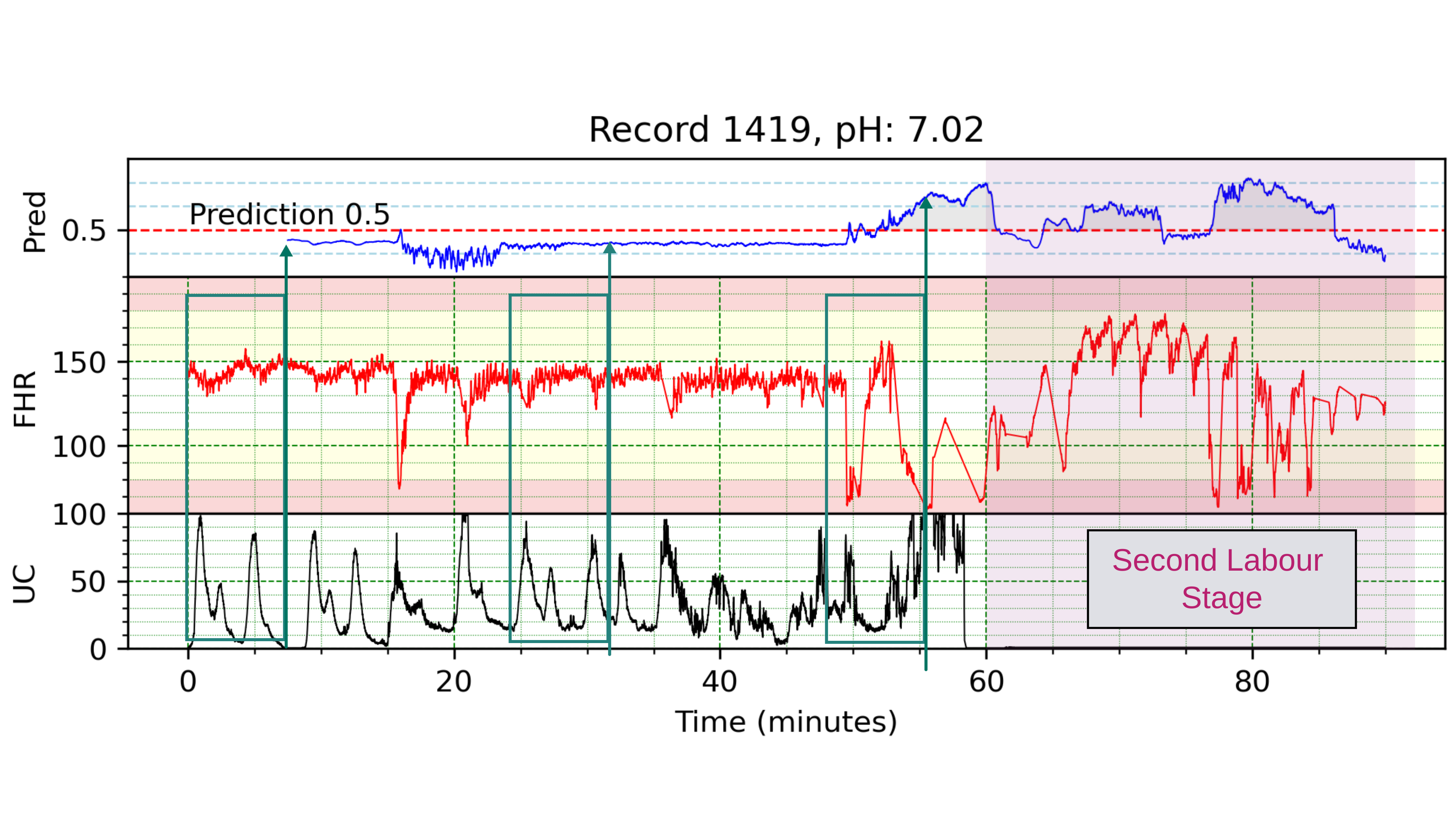}
\caption{Sliding window inference and alert generation. Example CTG recording with acidemic outcome (pH = 7.02). Top: Model prediction over time, with the 0.5 threshold indicated by the dashed red line. Light gray shading marks regions where predictions exceed the threshold, representing potential alerts. Cyan rectangles illustrate the 7.5-minute sliding windows used to compute each prediction point. Middle: Fetal heart rate (FHR) signal with background color bands indicating normal (green/yellow) and abnormal (pink) ranges. Bottom: Uterine contractions (UC). The shaded region on the right indicates the second stage of labor.}
\label{fig:inference}
\end{figure}

A key consideration for clinical deployment is that the model must operate in real-time during labor, rather than classifying entire CTG recordings retrospectively. To address this, we developed a sliding window inference approach that generates continuous risk predictions throughout the monitoring period (Figure~\ref{fig:inference}).

During inference, the model processes successive overlapping windows of 1800 samples (7.5 minutes at 4 Hz) across the entire CTG recording, producing a prediction score at each time point. Regions where the prediction exceeds 0.5 are identified as potential alerts, representing periods of suspected fetal compromise.

For each contiguous alert segment, we extracted summary statistics: segment length (duration above threshold), maximum prediction value, cumulative sum of predictions, and a weighted integral defined as $\sum_{t}(p_t - 0.5)^2$ for predictions $p_t$ in the segment. This weighted measure emphasizes sustained high-confidence predictions over brief threshold crossings.

To translate these segment-level features into clinically actionable alerts, we applied logistic regression using the largest alert segment from each recording. The features included segment length, maximum value, cumulative sum, and weighted integral. This two-stage approach—neural network for temporal pattern recognition followed by interpretable logistic regression for alert classification—provides both the representational capacity of deep learning and the transparency required for clinical decision support.

\section{Results}

Table~\ref{tab:results} summarizes classification performance on the held-out test set. The model achieved an AUC of 0.83 and accuracy of 79\% across all 55 test recordings (11 acidemia cases, 20.0\% prevalence). Excluding cesarean deliveries or abnormal presentations each improved performance to AUC of 0.85, with similar results when applying both criteria (n=46, AUC 0.85, accuracy 80\%).

\begin{table}[t]
\centering
\caption{Classification performance for fetal acidemia (pH < 7.15) on the test set.}
\label{tab:results}
\begin{tabular}{lccccc}
\hline
\textbf{Subgroup} & \textbf{N} & \textbf{Acidemia} & \textbf{Prevalence} & \textbf{AUC} & \textbf{Accuracy} \\
\hline
All test cases & 55 & 12 & 21.4\% & 0.826 & 0.786 \\
Vaginal delivery & 50 & 10 & 20.0\% & 0.850 & 0.800 \\
Cephalic presentation & 50 & 10 & 20.0\% & 0.848 & 0.800 \\
Vaginal + Cephalic & 46 & 9 & 19.6\% & 0.853 & 0.804 \\
No labor arrest & 47 & 8 & 17.0\% & 0.837 & 0.830 \\
Vaginal + Cephalic + No arrest & 43 & 7 & 16.3\% & 0.837 & 0.837 \\
\hline
\end{tabular}
\end{table}

\begin{figure}[t]
\centering
\includegraphics[width=\textwidth]{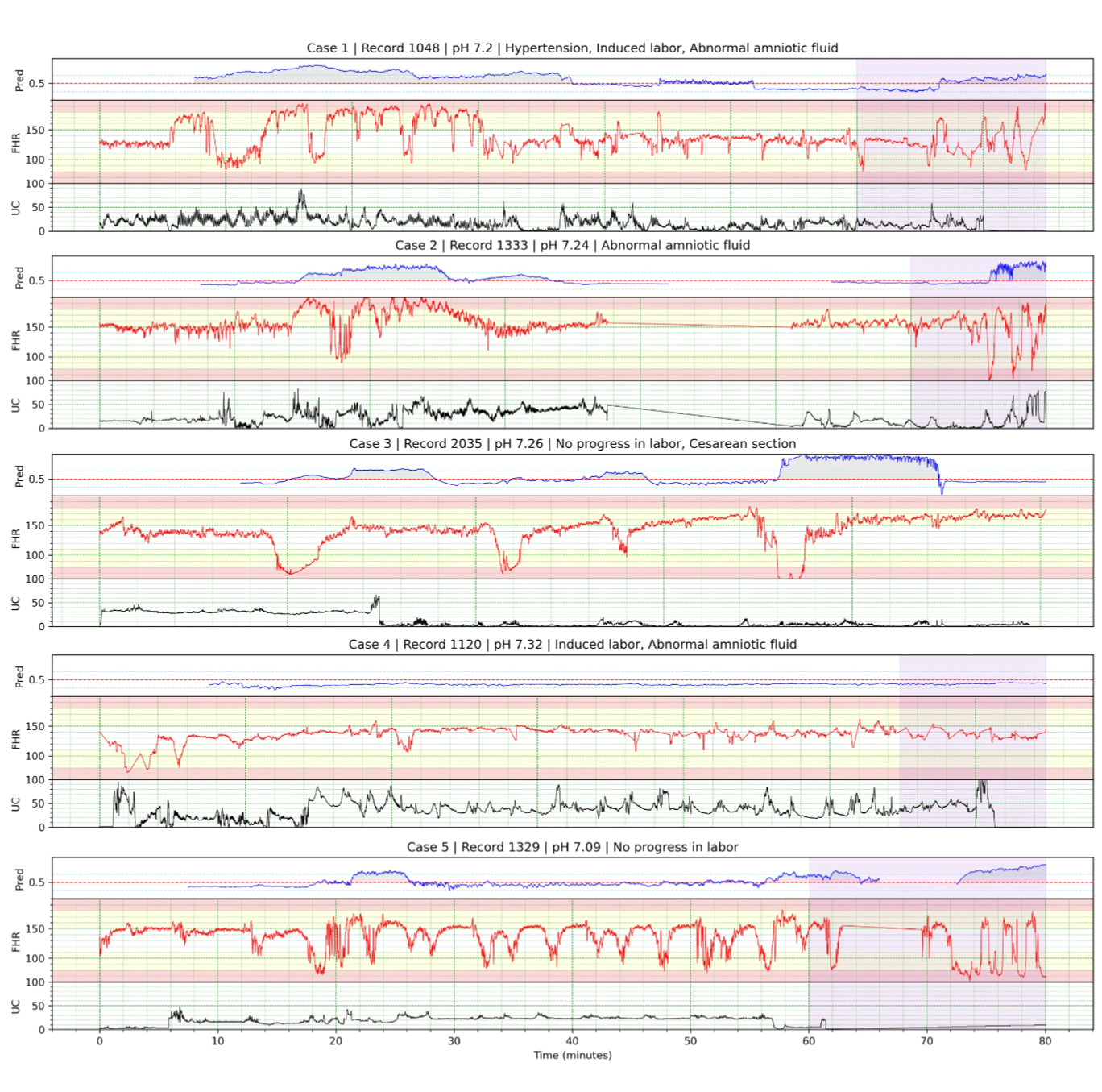}
\caption{Error analysis of alert generation. Each panel shows prediction score (top), FHR (middle), and UC (bottom). Cases 1--2: The two largest false positives, with abnormal CTG patterns despite normal pH. Case 3: Cesarean delivery with rising predictions before intervention. Case 4: True negative. Case 5: True positive with clear alert detecting acidemia (pH 7.09).}
\label{fig:examples}
\end{figure}

Figure~\ref{fig:examples} presents an error analysis of the alert-generation pipeline. Cases 1--2 are the two largest classification errors: both received the highest false-positive probabilities (0.58 and 0.32) yet have normal pH (7.20 and 7.24). However, both recordings show abnormal CTG patterns, and both labors involved risk factors (Case 1: hypertension, induced labor, abnormal amniotic fluid; Case 2: abnormal amniotic fluid). Case 3 illustrates the cesarean paradox: a cesarean delivery with no second stage of labor and normal pH (7.26), yet rising predictions—if the cesarean was performed due to CTG abnormalities, alerts would be appropriate even though the outcome was normal. Case 4 shows a true negative (pH 7.32) with no alerts, while Case 5 (pH 7.09) shows a clear alert before the second stage correctly identifying acidemia.

\section{Conclusion}

We presented the first application of self-supervised pre-training to intrapartum CTG analysis, achieving AUC of 0.853 on uncomplicated vaginal deliveries, and 0.83 AUC on the whole test set—the highest reported performance on the CTU-UHB benchmark. By leveraging 2,444 hours of unlabeled CTG recordings for masked pre-training, our approach demonstrates that the foundation model paradigm can help address data scarcity that has limited previous supervised methods. Error analysis revealed that most false-positive alerts corresponded to alarming CTG patterns confirmed by clinical review, suggesting the model captures clinically meaningful features.

Several limitations warrant consideration. The labeled CTU-UHB dataset originates from a single center (University Hospital Brno), while pre-training data spans three sources: CTU-UHB, FHRMA (Lille Catholic Hospital, France), and SPAM \cite{spam2017challenge,georgieva2019computer} (multi-center: Oxford, Lyon, Brno). Although this provides some diversity, signal characteristics vary across monitors and institutions, and broader representation is needed for robust generalization. The model processes CTG signals alone and does not incorporate clinical context such as maternal risk factors or labor interventions. For example, induced labor—present in 39\% of recordings—alters CTG patterns through stronger, more frequent contractions and increased risk of fetal heart rate abnormalities, information that would help the model contextualize observed patterns. The small dataset size precluded including such features.

The main limitation remains availability of labeled data at scales suitable for modern transformer models. Future directions include longer input windows, which the patch-based architecture supports efficiently since longer signals embed into the same number of patches. Clinical context should be incorporated when larger datasets permit.

Privacy constraints limit direct sharing of clinical CTG data. However, we release our model weights and standardized benchmark splits to enable reproducibility. Clinical centers can continue pre-training on their local recordings and share only updated weights, expanding model diversity while preserving patient privacy. In the labor room, online predictions on the specific recording in progress could enable personalized alerts generation.

Despite current limitations, our results demonstrate that foundation models can extract meaningful representations from CTG signals even with limited labeled data. We release our code and trained weights to facilitate further research.

\section*{Code and Data Availability}

The CTGDL dataset integrating CTU-UHB, FHRMA, and SPaM recordings is available on Zenodo at \url{https://doi.org/10.5281/zenodo.18034361} \cite{fridman2025ctgdl}. The repository includes standardized benchmark splits (train/validation/test) to enable reproducible evaluation. Pre-trained model weights and training code are available at \url{https://github.com/[repository]}.

\clearpage
\bibliography{mybibfile}

@article{chudacek2014ctuuhb,
  title        = {Open access intrapartum {CTG} database},
  author       = {Chud{\'a}{\v{c}}ek, V. and {\v{S}}pl{\'a}tek, J. and Bur{\v{s}}a, M. and Jank{\r{u}}, P. and Hruban, L. and Huptych, M. and Zach, L. and Lhotsk{\'a}, L.},
  journal      = {BMC Pregnancy and Childbirth},
  volume       = {14},
  pages        = {16},
  year         = {2014},
  doi          = {10.1186/1471-2393-14-16}
}

@article{boudet2019fhrma,
  title        = {Fetal heart rate signal dataset for training morphological analysis methods and evaluating them against an expert consensus},
  author       = {Boudet, S. and Houz{\'e} de l'Aulnoit, A. and Demailly, R. and Delgranche, A. and Peyrodie, L. and Beuscart, R. and Houz{\'e} de l'Aulnoit, D.},
  journal      = {Data in Brief},
  volume       = {27},
  pages        = {104720},
  year         = {2019},
  doi          = {10.1016/j.dib.2019.104720}
}

@article{chudacek2014open,
  title={Open access intrapartum {CTG} database},
  author={Chud{\'a}{\v{c}}ek, V{\'a}clav and Spilka, Ji{\v{r}}{\'\i} and Bur{\v{s}}a, Miroslav and Jank{\r{u}}, Petr and Hruban, Luk{\'a}{\v{s}} and Huptych, Michal and Lhotsk{\'a}, Lenka},
  journal={BMC Pregnancy and Childbirth},
  volume={14},
  number={1},
  pages={16},
  year={2014},
  publisher={BioMed Central},
  doi={10.1186/1471-2393-14-16}
}

@article{physionet2000,
  title={{PhysioBank, PhysioToolkit, and PhysioNet}: Components of a new research resource for complex physiologic signals},
  author={Goldberger, Ary L and Amaral, Luis AN and Glass, Leon and Hausdorff, Jeffrey M and Ivanov, Plamen Ch and Mark, Roger G and Mietus, Joseph E and Moody, George B and Peng, Chung-Kang and Stanley, H Eugene},
  journal={Circulation},
  volume={101},
  number={23},
  pages={e215--e220},
  year={2000},
  publisher={American Heart Association}
}

@misc{ctgdl2025,
  title={{CTGDL}: A Multi-source Cardiotocography Dataset for Fetal Stress Prediction and {CTG} Analysis},
  author={[Authors]},
  year={2025},
  publisher={Zenodo},
  doi={10.5281/zenodo.17903226}
}

@report{unigme2023,
  title={Never Forgotten: The Situation of Stillbirth Around the Globe},
  author={{United Nations Inter-agency Group for Child Mortality Estimation (UN IGME)}},
  institution={United Nations Children's Fund},
  address={New York},
  year={2023}
}

@article{vogel2014,
  title={Maternal complications and perinatal mortality: findings of the {World Health Organization Multicountry Survey on Maternal and Newborn Health}},
  author={Vogel, Joshua P and others},
  journal={BJOG: An International Journal of Obstetrics \& Gynaecology},
  volume={121},
  number={Suppl 1},
  pages={76--88},
  year={2014}
}

@article{bhutta2014,
  title={Can available interventions end preventable deaths in mothers, newborn babies, and stillbirths, and at what cost?},
  author={Bhutta, Zulfiqar A and others},
  journal={The Lancet},
  volume={384},
  number={9940},
  pages={347--370},
  year={2014}
}

@article{goldenberg2016,
  title={Stillbirths: The Hidden Birth Asphyxia -- {US} and Global Perspectives},
  author={Goldenberg, Robert L and Harrison, Michelle S and McClure, Elizabeth M},
  journal={Clinics in Perinatology},
  volume={43},
  number={3},
  pages={439--453},
  year={2016}
}

@article{bennet2009,
  title={The fetal heart rate response to hypoxia: insights from animal models},
  author={Bennet, Laura and Gunn, Alistair J},
  journal={Clinics in Perinatology},
  volume={36},
  number={3},
  pages={655--672},
  year={2009}
}

@article{ayres2015figo,
  title={{FIGO} consensus guidelines on intrapartum fetal monitoring: Physiology of fetal oxygenation and the main goals of intrapartum fetal monitoring},
  author={Ayres-de-Campos, Diogo and Arulkumaran, Sabaratnam and {FIGO Intrapartum Fetal Monitoring Expert Consensus Panel}},
  journal={International Journal of Gynecology \& Obstetrics},
  volume={131},
  number={1},
  pages={5--8},
  year={2015}
}

@article{alfirevic2017,
  title={Continuous cardiotocography ({CTG}) as a form of electronic fetal monitoring ({EFM}) for fetal assessment during labour},
  author={Alfirevic, Zarko and Gyte, Gillian ML and Cuthbert, Anna and Devane, Declan},
  journal={Cochrane Database of Systematic Reviews},
  number={2},
  year={2017},
  publisher={John Wiley \& Sons, Ltd}
}

@article{bernardes1997,
  title={Evaluation of interobserver agreement of cardiotocograms},
  author={Bernardes, Jo{\~a}o and Costa-Pereira, Altamiro and Ayres-de-Campos, Diogo and van Geijn, Herman P and Pereira-Leite, Lu{\'\i}s},
  journal={International Journal of Gynecology \& Obstetrics},
  volume={57},
  number={1},
  pages={33--37},
  year={1997}
}

@article{palomaki2006,
  title={Intrapartum cardiotocography -- the dilemma of interpretational variation},
  author={Palom{\"a}ki, Outi and Luukkaala, Tiina and Luoto, Riitta and Tuimala, Risto},
  journal={Journal of Perinatal Medicine},
  volume={34},
  number={4},
  pages={298--302},
  year={2006}
}

@article{schiermeier2011,
  title={Intra- and interobserver variability of intrapartum cardiotocography: a multicenter study comparing the {FIGO} classification with computer analysis software},
  author={Schiermeier, Sven and Westhof, Gert and Leven, Alexandra and Hatzmann, Heinrich and Reinhard, J{\"u}rgen},
  journal={Gynecologic and Obstetric Investigation},
  volume={72},
  number={3},
  pages={169--173},
  year={2011}
}

@article{blencowe2016,
  title={National, regional, and worldwide estimates of stillbirth rates in 2015, with trends from 2000: a systematic analysis},
  author={Blencowe, Hannah and others},
  journal={The Lancet Global Health},
  volume={4},
  number={2},
  pages={e98--e108},
  year={2016}
}

@article{lawn2016,
  title={Stillbirths: rates, risk factors, and acceleration towards 2030},
  author={Lawn, Joy E and others},
  journal={The Lancet},
  volume={387},
  number={10018},
  pages={587--603},
  year={2016}
}

@article{ayres2011bias,
  title={Knowledge of adverse neonatal outcome alters clinicians' interpretation of the intrapartum cardiotocograph},
  author={Ayres-de-Campos, Diogo and Arteiro, Daniela and Costa-Santos, Cristina and Bernardes, Jo{\~a}o},
  journal={BJOG: An International Journal of Obstetrics \& Gynaecology},
  volume={118},
  number={8},
  pages={978--984},
  year={2011}
}

@inproceedings{warrick2010,
  title={A Machine Learning Approach to the Detection of Fetal Hypoxia during Labor and Delivery},
  author={Warrick, Philip and Hamilton, Emily and Kearney, Robert and Precup, Doina},
  booktitle={Proceedings of the AAAI Conference on Artificial Intelligence},
  volume={24},
  number={1},
  pages={1865--1870},
  year={2010}
}

@article{warrick2012,
  title={A Machine Learning Approach to the Detection of Fetal Hypoxia during Labor and Delivery},
  author={Warrick, Philip and Hamilton, Emily and Kearney, Robert and Precup, Doina},
  journal={AI Magazine},
  volume={33},
  number={2},
  pages={79},
  year={2012}
}

@article{hoodbhoy2019,
  title={Use of machine learning algorithms for prediction of fetal risk using cardiotocographic data},
  author={Hoodbhoy, Zahra and others},
  journal={International Journal of Applied and Basic Medical Research},
  volume={9},
  number={4},
  pages={226},
  year={2019}
}

@inproceedings{pradhan2021,
  title={A Machine Learning Approach for the Prediction of Fetal Health using {CTG}},
  author={Pradhan, Ashis Kumar and Rout, Jitendra Kumar and Maharana, Ajay Bhuyan and Balabantaray, Bunil Kumar and Ray, Niranjan Kumar},
  booktitle={2021 19th OITS International Conference on Information Technology (OCIT)},
  pages={239--244},
  year={2021},
  organization={IEEE}
}

@article{sisporto2000,
  title={{SisPorto 2.0}: a program for automated analysis of cardiotocograms},
  author={Ayres-de-Campos, Diogo and Bernardes, Jo{\~a}o and Garrido, Ant{\'o}nio and Marques-de-S{\'a}, Joaquim and Pereira-Leite, Lu{\'\i}s},
  journal={Journal of Maternal-Fetal Medicine},
  volume={9},
  number={5},
  pages={311--318},
  year={2000}
}

@article{fergus2020,
  title={Modelling Segmented Cardiotocography Time-Series Signals Using One-Dimensional Convolutional Neural Networks for the Early Detection of Abnormal Birth Outcomes},
  author={Fergus, Paul and others},
  journal={IEEE Transactions on Emerging Topics in Computational Intelligence},
  volume={5},
  number={6},
  pages={882--892},
  year={2020}
}

@article{zhao2019,
  title={{DeepFHR}: intelligent prediction of fetal Acidemia using fetal heart rate signals based on convolutional neural network},
  author={Zhao, Zhidong and others},
  journal={BMC Medical Informatics and Decision Making},
  volume={19},
  number={1},
  pages={286},
  year={2019}
}

@article{kalyan2021,
  title={{AMMUS}: A Survey of Transformer-based Pretrained Models in Natural Language Processing},
  author={Kalyan, Katikapalli Subramanyam and Rajasekharan, Ajit and Sangeetha, Sivanesan},
  journal={arXiv preprint arXiv:2108.05542},
  year={2021}
}

@article{khan2021,
  title={Transformers in Vision: A Survey},
  author={Khan, Salman and others},
  journal={ACM Computing Surveys},
  volume={54},
  number={10s},
  pages={1--41},
  year={2022}
}

@inproceedings{karita2019,
  title={A Comparative Study on Transformer vs {RNN} in Speech Applications},
  author={Karita, Shigeki and others},
  booktitle={IEEE Automatic Speech Recognition and Understanding Workshop (ASRU)},
  pages={449--456},
  year={2019}
}

@article{patchtst2023,
  title={A Time Series is Worth 64 Words: Long-term Forecasting with Transformers},
  author={Nie, Yuqi and Nguyen, Nam H and Sinthong, Phanwadee and Kalagnanam, Jayant},
  journal={arXiv preprint arXiv:2211.14730},
  year={2023}
}

@article{chudacek2014,
  title={Open access intrapartum {CTG} database},
  author={Chud{\'a}{\v{c}}ek, V{\'a}clav and Spilka, Ji{\v{r}}{\'\i} and Bur{\v{s}}a, Miroslav and Jank{\r{u}}, Petr and Hruban, Luk{\'a}{\v{s}} and Huptych, Michal and Lhotsk{\'a}, Lenka},
  journal={BMC Pregnancy and Childbirth},
  volume={14},
  number={1},
  pages={16},
  year={2014},
  doi={10.1186/1471-2393-14-16}
}

@article{ogasawara2021,
  title={Deep neural network-based classification of cardiotocograms outperformed conventional algorithms},
  author={Ogasawara, Jun and others},
  journal={Scientific Reports},
  volume={11},
  number={1},
  pages={13367},
  year={2021}
}

@article{petrozziello2019,
  title={Multimodal convolutional neural networks to detect fetal compromise during labor and delivery},
  author={Petrozziello, Alessio and Redman, Christopher WG and Papageorghiou, Aris T and Jordanov, Ivan and Georgieva, Antoniya},
  journal={IEEE Access},
  volume={7},
  pages={112026--112036},
  year={2019}
}

@article{mbarek2023,
  title={{DeepCTG} 1.0: an interpretable model to detect fetal hypoxia from cardiotocography data during labor and delivery},
  author={Ben M'Barek, Imane and Jauvion, Gr{\'e}goire and Vitrou, Jade and others},
  journal={Frontiers in Pediatrics},
  volume={11},
  pages={1190441},
  year={2023}
}

@article{khan2024patchctg,
  title={{PatchCTG}: Patch Cardiotocography Transformer for Antepartum Fetal Health Monitoring},
  author={Khan, Muhammad Jaleed and Vatish, Manu and Davis Jones, Gabriel},
  journal={arXiv preprint arXiv:2411.07796},
  year={2024}
}

@article{wu2024etcnn,
  title={{ETCNN}: An ensemble transformer-convolutional neural network for automatic analysis of fetal heart rate},
  author={Wu, Qingjian and Lu, Yaosheng and Kang, Xue and Wang, Huijin and Zheng, Zheng and Bai, Jieyun},
  journal={Biomedical Signal Processing and Control},
  volume={96},
  pages={106629},
  year={2024},
  publisher={Elsevier}
}

@inproceedings{spam2017challenge,
  title        = {CTG Challenge: SPaM in Labour Workshop 2017 Intrapartum Cardiotocography Database},
  booktitle    = {SPaM in Labour Workshop, Oxford},
  year         = {2017},
  note         = {\url{https://users.ox.ac.uk/~ndog0178/CTGchallenge2017.htm}}
}

@article{georgieva2019computer,
  title   = {Computer-based intrapartum fetal monitoring and beyond},
  author  = {Georgieva, A. and Payne, S. J. and Moulden, M. and Redman, C. W. G.},
  journal = {Acta Obstetricia et Gynecologica Scandinavica},
  year    = {2019},
  volume  = {98},
  number  = {5},
  pages   = {550--559},
  doi     = {10.1111/aogs.13639}
}

@dataset{fridman2025ctgdl,
  author       = {Fridman, Naomi},
  title        = {{CTGDL}: A Multi-source Cardiotocography Dataset for Fetal Stress Prediction and {CTG} Analysis},
  month        = dec,
  year         = 2025,
  publisher    = {Zenodo},
  doi          = {10.5281/zenodo.18034361},
  url          = {https://doi.org/10.5281/zenodo.18034361}
}

\end{document}